\documentclass[conference]{IEEEtran}
\usepackage{cite}

\ifCLASSINFOpdf
   \usepackage[pdftex]{graphicx}
\else
   \usepackage[dvips]{graphicx}
\fi
\usepackage{amsmath}
\usepackage{algpseudocode}

\usepackage{array}
\usepackage[bookmarks=false]{hyperref}

\begin{document}
%
\title{Data Agnostic RoBERTa-based Natural Language to SQL Query Generation}

\author{\IEEEauthorblockN{Debaditya Pal}
\IEEEauthorblockA{Student of \\ Computer Science and Engineering \\
ABV-IIITM, Gwalior\\
Gwalior, Madhya Pradesh 474015\\
Email: debaditya.pal6@gmail.com}
\and
\IEEEauthorblockN{Harsh Sharma}
\IEEEauthorblockA{Student of \\ Computer Science and Engineering \\
ABV-IIITM, Gwalior\\
Gwalior, Madhya Pradesh 474015 \\
Email: harshhsharma23@gmail.com}
\and
\IEEEauthorblockN{Kaustubh Chaudhuri}
\IEEEauthorblockA{Student of \\ Computer Science and Engineering \\
ABV-IIITM, Gwalior\\
Gwalior, Madhya Pradesh 474015 \\
Email: ckaustubhm06@gmail.com}
}


%


\IEEEoverridecommandlockouts
\IEEEpubid{\makebox[\columnwidth]{978-1-7281-8876-8/21/\$31.00 ©2021 IEEE \hfill} \hspace{\columnsep}\makebox[\columnwidth]{ }}
\maketitle
\IEEEpubidadjcol
\begin{abstract}
Relational databases are among the most widely used architectures to store massive amounts of data in the modern world. However, there is a barrier between these databases and the average user. The user often lacks the knowledge of a query language such as SQL required to interact with the database. The NL2SQL task aims at finding deep learning approaches to solve this problem by converting natural language questions into valid SQL queries. Given the sensitive nature of some databases and the growing need for data privacy, we have presented an approach with data privacy at its core. We have passed RoBERTa embeddings and data-agnostic knowledge vectors into LSTM based submodels to predict the final query. Although we have not achieved state of the art results, we have eliminated the need for the table data, right from the training of the model, and have achieved a test set execution accuracy of 76.7\%. By eliminating the table data dependency while training we have created a model capable of zero shot learning based on the natural language question and table schema alone.
\end{abstract}


%
\IEEEpeerreviewmaketitle

\section{Introduction}
Human beings have been exponentially generating vast amounts of data over the past few decades, be it medical records or even financial market-related information \cite{FinancialRec}. Relational databases have proven to be a highly reliable database architecture to store this data due to their simplicity and intuitiveness. However, accessing data through these databases necessitates a user to have prerequisite knowledge of a query language such as SQL. This creates a barrier between the typical non-technical user and the benefits of these databases. This paved the way for the NL2SQL task which is aimed at improving human-computer interaction by eliminating the need for the user to know a query language. Solving the task involves coming up with deep learning approaches to convert a natural language question into an SQL query. The state of the art models \cite{IESQL} can achieve up to 87.8\% logical form accuracy in these queries but most of the high accuracy models depend on the table data in some manner, either by using them as features while training, or by checking query results to perform execution guided decoding to achieve such high levels of accuracy.
Most modern databases store sensitive and private records and hence, data privacy has become of paramount importance. Owing to the large size of these databases, they are often distributed among different servers. In order to get predictions through a conventional NL2SQL model, the entirety of the data needs to be transferred, using up unnecessary bandwidth and making the data susceptible to man-in-the-middle attacks. Moreover using these models as 3$^{rd}$ party services requires the user to entrust the 3$^{rd}$ party with all of their data, without the guarantee of the data not being used for other purposes without consent. Keeping this in mind, we introduce our approach that utilizes only the table schema and is entirely independent of the table data at each step of its operation, thus, ensuring data privacy by eliminating unnecessary flow of data. Our main contributions to the field are as follows:
\begin{enumerate}
\item {We eliminate the utilization of the table data, creating a data-agnostic model that gives predictions based on the table schema and the natural language question alone.}
\item {We create a model capable of zero shot learning. As the model does not explicitly train on one particular table by using its cell data, we can make a highly scalable model that can predict queries on previously unseen databases.}
\item {We employ a transformer model named RoBERTa to create embeddings from the natural language question and the table schema which serves as a latent space representation of the data. The final prediction model uses these embeddings along with two knowledge vectors to come up with the final prediction.}
\end{enumerate}

\section{Related Work}
The dataset used was the WikiSQL dataset \cite{SEQ2SQL} , a corpus of 80654 hand-annotated instances of natural language questions, SQL queries, and SQL tables extracted from 24,241 HTML tables from Wikipedia. It can be broadly classified as a semantic parsing database.
BERT \cite{BERT}  is a deep bidirectional transformer based model that is first pre-trained on a very large text corpus using masked language model loss and then next sentence prediction loss. This gives the model some “knowledge” or “context” of the language it is being trained on. The concept of using a transformer such as BERT as an encoder to generate feature embeddings was inspired by the “A Comprehensive Exploration on WikiSQL with Table-Aware Word Contextualization” paper \cite{SQLova}.
We have used a BERT-based model named RoBERTa \cite{RoBERTa}, which uses a more optimized pretraining method and has been proven to perform better than its predecessors. RoBERTa uses dynamic masking during its pretraining on the masked language model task, replaces the next sentence prediction task, and uses a much bigger text corpus in general to achieve better results.
The paper titled “Content Enhanced BERT-based Text-to-SQL Generation” \cite{GuoTong} introduces two table content-aware vectors, viz. the question mark and header mark vectors, that can be used as features to improve the performance of the overall model. Henceforth, the term "header" shall refer to the names of the table columns. We have built on top of this knowledge and come up with a different algorithm to extract such vector features, without using the table content. The motivation behind using these knowledge vectors is to provide the model with highly correlated words between the natural language question and the table schema which serves as additional features to aid the prediction task. 

\hypertarget{III}{\section{Generation of Knowledge Vectors}} 
As described above, we have introduced two knowledge vectors, the Question Mark Vector and the Header Mark Vector. These knowledge vectors are boolean vectors which encode the knowledge of which headers and which tokens in the question may be of significant importance to the model by virtue of correlation. These vectors are first concatenated together, and then passed to the model as additional features. A visual representation of this process may be found in \hyperlink{Fig:1}{Fig 1}.

\subsection{Generation of Question Mark Vector}
\medskip

\begin{algorithmic}
\State $vector$ = [0] * len($natural\_language\_question$)
\For{$word$ in $natural\_language\_question$}
    \If {contain($headers$, $word$)}
        \State $index$ = get\_index($word$)
        \State $vector[index]$ = 1
    \EndIf
\EndFor
\end{algorithmic}
\medskip
The function “contain” is described below:
\begin{algorithmic}
\Function {contain}{$sentence, phrase$}
    \For{$word$ in $sentence$}
        \If{$word == phrase$}
        \State \Return $true$
        \EndIf
    \EndFor {}
    \State \Return $false$
\EndFunction
\end{algorithmic}

If the given phrase matches any of the words in the sentence completely (full match), it returns $true$. If the phrase does not completely match with any of the words in the sentence, a value of $false$ is returned.

Applying this function to the headers, we check for each word in the natural language question, whether that word is contained within the headers or not. If the contain function returns a value of $true$, we mark the index where the word is located with the value “1” to denote that this word may be of importance to the model when it comes to generating the SQL query. Otherwise, the value of the vector remains “0”.

\subsection{Algorithm: Generation of Header Mark Vector}
\medskip

\begin{algorithmic}
\State $vector$ = [0] * len($headers$) 
\For{$word$ in $headers$}
    \If {contain($natural\_language\_question$, $word$)}
        \State $index$ = get\_index($word$)
        \State $vector[index]$ = 1
    \EndIf
\EndFor
\end{algorithmic}
\medskip
Note that here we iterate not over the headers themselves, but over every word in the header as well. This allows us to check for partial matches from the natural language question in the list of headers as well.

Applying the contain function to the natural language question this time, we check for each word in the list of headers, whether that word is contained within the natural language question or not. It is our observation that often when a word in the table schema and the natural language question match, the header is a part of the query. The table cell content has not been utilized in the generation of these vectors, which is in line with our goal of data privacy. 

For brevity, we shall henceforth refer to the Question Mark Vector as QMV and the Header Mark Vector as HMV.

\section{The Model}
A single SQL query might have multiple equally valid serializations. We cannot train a sequence-to-sequence-style model, as it chooses one correct serialization and penalizes the algorithm for picking other valid serializations. This is a problem documented as the “order matters” problem. Hence we have decided to follow a sketch-based approach as described in the SQLNet paper. The basic structure of every SQL query is the same, which reduces the task of generating complete SQL queries to predicting only certain key characteristics of the SQL query, much like a fill-in-the-blanks task. The relative ordering of the where clauses inside a sketch do not matter and the model is not penalized for choosing an order that is different from the ground truth query as long as all the clauses have been correctly predicted. The following is the sketch of an SQL query:
\begin{center}
{\bf SELECT} \{$aggregate$\} \{$column$\} {\bf FROM} \{$table$\} \newline
{\bf WHERE} \{$column$\} \{$operator$\} \{$value$\}\newline
{\bf AND*} \newline
{\bf WHERE} \{$column$\} \{$operator$\} \{$value$\}\newline
*(Repeating {\bf WHERE} blocks)
\end{center}

For our model, the initial step is to generate RoBERTa Embeddings from the natural language question and the table schema. We then generate the two knowledge vectors with the algorithms described above. The next step involves passing these features into 6 sub-models to generate the values which will be filled into the SQL sketch for the final SQL query.

\begin{figure}[h]
\hypertarget{Fig:1}{}
\centerline{\includegraphics[width=18.5pc]{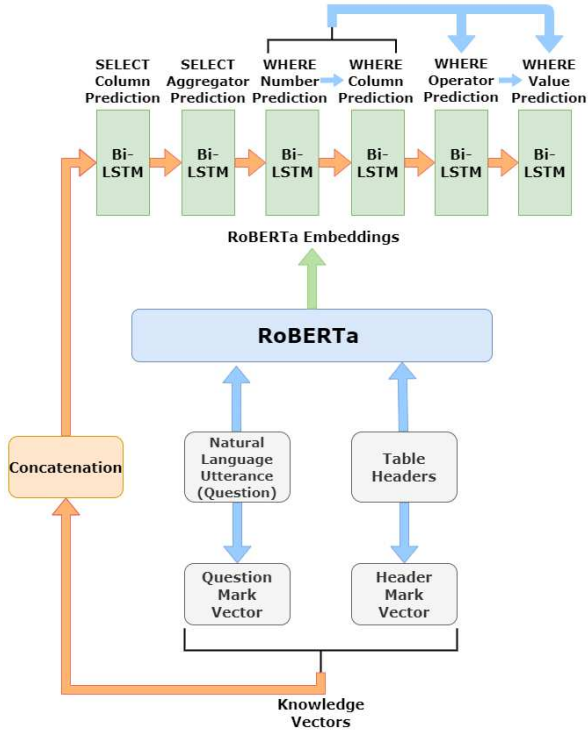}}
\caption{Architecture of the model showing the different layers }
\end{figure}
\subsection{RoBERTa Embedding Layer}
The NL2SQL task is essentially a semantic parsing task and we have observed that the tokens of the natural language question might not always match with the tokens in the header, thus sometimes, the two knowledge vectors may fail to provide the downstream submodels with useful information. We must try to extract a significant amount of information from the natural language question and the table schema. We have chosen to use RoBERTa as the base model in our architecture. We have used tokenized input to keep a track of the indices of possible “Where Value” (i.e. the value(s) to be filled in the where clause of the aforementioned sketch) subsequences.
RoBERTa’s default tokenizer preprocesses the data by adding sentinel tokens into the input sequence and then converting all the tokens into their respective IDs from the vocabulary it was trained on.
We generate two embeddings from this layer; one to represent the context of the natural language question, and the other to represent the context of the table schema in downstream tasks.

\subsection{Select Aggregate Prediction}
\begin{figure}[!t]
\hypertarget{Fig:2}{}
\centering
\includegraphics[width=3in]{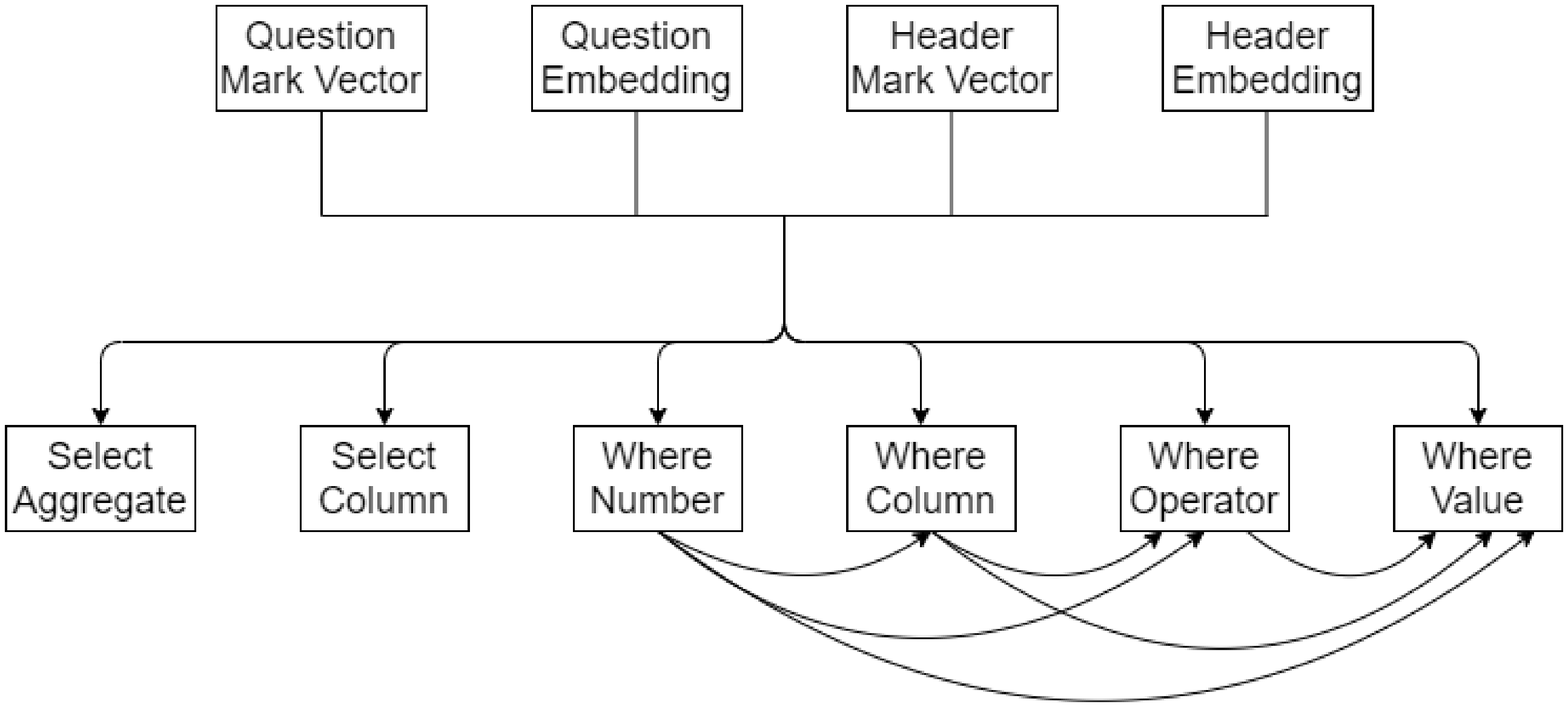}
\caption{A Dependency Graph showing the inputs needed for the respective predictions}
\label{dependency_graph}
\end{figure}
The Select Aggregate performs an operation on multiple data points and then returns a single value, for eg: Average, Minimum, Maximum, etc. Its main purpose is to calculate the general trend over multiple instances of data. The problem has been formulated as:
\begin{equation}
    P(sa) = f_{1,sa}(f_{2,sa}(Q,QMV,H,HMV) + b_{sa})
\end{equation}
Where $f_{1,sa}$ and $f_{2,sa}$ are functions learnt by the model and $b_{sa}$ is the bias.

We have passed the above features into an bidirectional LSTM based network and formulated the problem as a classification task where the aggregates are the different classes to obtain a list of probabilities.

\subsection{Select Column Prediction}
The Select Column refers to the column from which the final displayed data is to be taken. We have limited ourselves to single column querying as the WikiSQL dataset does not have the required data to train a multi-column query model. We can observe that the Select Column is dependent on both the natural language question as well as the table schema. The natural language question affects the probability of a column being the Select Column while the table schema provides us with the different columns in the table. The problem has been formulated as:
\begin{equation}
    P(sc) = f_{1,sc}(f_{2,sc}(Q,QMV,H,HMV) + b_{sc})
\end{equation}
Where $f_{1,sc}$ and $f_{2,sc}$ are functions learnt by the model and $b_{sc}$ is the bias.
 Even if the natural language question does not contain the name of a header specifically, the RoBERTa embedding of the word used to indicate the header will be close value to the desired value as the two words have been used in the same context. We have made an important assumption here that the column names are descriptive of the column content. 

\subsection{Where Number Prediction}
This value isn’t directly present in the final predicted SQL query, however, it is responsible for the structure of the query itself as it determines the number of where clauses in the query. We can observe that the Where Number is dependent on both the natural language question as well as the table schema. The problem has been formulated as:

\begin{equation}
    P(wn) = f_{1,wn}(f_{2,wn}(Q,QMV,H,HMV) + b_{wn})
\end{equation}
Where $f_{1,wn}$ and $f_{2,wn}$ are functions learnt by the model and $b_{wn}$ is the bias.
The natural language question might contain conjunctions that directly point us towards the number of where clauses. The table schema might help the model decide the Where Number, in the rare case that a phrase in the natural language question might refer to multiple columns in the table schema.

\subsection{Where Column Prediction}
The Where Column describes the column with which the Where Value is to be compared with. The natural language question guides the selection of the Where Column, the table schema provides the list from which the column is to be selected. It is important to note that the prediction of the Where Column also depends on the Where Number as that value serves as the number of predictions the model needs to output. The problem has been formulated as:
\begin{equation}
    P(wc) = f_{1,wc}(f_{2,wc}(Q,QMV,H,HMV,P_{wn}) + b_{wc})
\end{equation}
Where $f_{1,wc}$ and $f_{2,wc}$ are functions learnt by the model and $b_{wc}$ is the bias.
 Since the components of the where clause are interlinked, we have used a beam search algorithm to compute the most probable combination of the Where Column, Operator, and Value. We have achieved an increment of approximately 5\% in logical form accuracy after using beam search as compared to the normal method.

\subsection{Where Operator Prediction}
It provides the description of how the Where Value is to be compared with the data in the queried Where Column. The problem has been formulated as:
\begin{equation}
    P(wo) = f_{1,wo}(f_{2,wo}(Q,QMV,H,HMV,P_{wn}, P_{wc}) + b_{wo})
\end{equation}
Where $f_{1,wo}$ and $f_{2,wo}$ are functions learnt by the model and $b_{wo}$ is the bias.
We can observe that the Where Operator is dependent on the natural language question itself and the type of the Where Column. Thus, the Where Column heavily influences the selection of the Where Operator.

\subsection{Where Value Prediction}
 It provides the data which is compared with the queried Where Column data. An important observation to be made is that the Where Value is always a subsequence of the natural language question. Thus if the model outputs the start and end index of the tokenized natural language question, we can easily form the Where Value phrase from that information. So we train the sub model to output two indices marking the start and end indices of the Where Value phrase present in the natural language question. We formulate the problem as:
\begin{equation}
\begin{split}
P_(wv) = f_{1,wv}(f_{2,wv}(Q,QMV,H,HMV,  \\
         P_{wn},P_{wc},P_{wo}) + b_{wv})
\end{split}
\end{equation}
Where $f_{1,wv}$ and $f_{2,wv}$ are functions learnt by the model and $b_{wv}$ is the bias.
We have used beam searching to optimize performance. Since we have decided to remove the table cell data, this sub-task has been adversely affected. In addition to being a subsequence of the tokenized natural language sequence, the Where Value, especially in cases when the Where Operator is ‘=’, will always be present inside the table. Thus using table data would give the model a huge boost in confidence for predicting a certain set of words as the desired Where Value. 

\section{Results}
We have evaluated our model on the WikiSQL dataset with the logical form and execution accuracy metrics. The method of calculating the execution accuracy does involve the model querying the database which goes against our core objective of data privacy, but this is not a necessary step after the model has been deployed and neither does it affect the model’s training in any way. The only purpose of such querying is to obtain an additional evaluation metric to better understand our model’s performance.
It is important to note that there exists a performance v/s privacy tradeoff, as ensuring data privacy also means giving less information to the model itself, we have tried to maximize our model's performance without compromising our core objective of data privacy. Nevertheless, We have compared our model with the state of the art model, latest content insensitive model:

\noindent Note: "Acc$_{lf}$" stands for logical form accuracy, whereas "Acc$_{ex}$" stands for execution accuracy.
\begin{center}
\begin{tabular}{|m{7em}|m{3em}|m{3em}|m{3em}|m{3em}| } 
 \hline
Model & Dev Acc$_{lf}$. & Dev Acc$_{ex}$. & Test Acc$_{lf}$. & Test Acc$_{ex}$. \\ 
\hline
IE-SQL(State of the Art)\cite{IESQL} & 87.9 & 92.6 & 87.8 & 92.5 \\
\hline
TypeSQL(content-insensitive)\cite{TypeSQL} & - & 74.5 & - & 73.5 \\
\hline
Our Model & 69.4 & 77.0 & 68.9 & 76.7 \\
\hline
\end{tabular}
\end{center}

The following are the accuracies for the individual tasks performed by the sub-models:
\begin{center}
\begin{tabular}{|m{8em}|m{4em}|m{4em}| } 
\hline
Dataset & Dev & Test \\
\hline
Select Aggregate & 90.4 & 90.3 \\
\hline
Select Column & 95.1 & 94.6 \\
\hline
Where Number & 97.7 & 97.7 \\
\hline
Where Column & 88.8 & 87.6 \\
\hline
Where Operator & 91.2 & 90.7 \\
\hline
Where Value & 85.1 & 84.7 \\
\hline
\end{tabular}
\end{center}

\section {Zero Shot Learning}
We prepared a previously unseen table set for the model by eliminating the intersection of the tables that were present in both train and test sets from the test table set. Then we took the queries based on these tables to obtain a dataset to test for the zero shot learning capabilities of the model and obtained the following metrics. These schemas were completely unknown to the model before running predictions on them.

\begin{center}
\begin{tabular}{|m{8em}|m{4em}| } 
\hline
Dataset & Zero Shot \\
\hline
Select Aggregate & 90.1 \\
\hline
Select Column & 94.9 \\
\hline
Where Number & 97.4 \\
\hline
Where Column & 87.0 \\
\hline
Where Operator & 89.5 \\
\hline
Where Value & 83.6 \\
\hline
\hline
Logical Form Acc & 66.3 \\
\hline
Execution Acc & 74.7 \\
\hline
\end{tabular}
\end{center}

Since our model does not use the cell data present in the tables while training and making predictions, it prevents the model from overfitting on the training tables. Most importantly, it results in a highly scalable generalized model. The similarity in the metrics obtained from the two experimental setups proves that the model is highly generalized and has no high variance problems.

\section{Performance v/s Privacy Tradeoff}
While data privacy might be desirable, it does have a detrimental impact on the overall performance of the model. The knowledge vectors are being given lesser information as the role of the tabular data in their generation is being completely removed. Hence the model has less information to base its predictions on. It could be the case sometimes that none of the words in the natural language question match with the table headers, but instead, they match with some cell in the table data. In such cases, our model has to predict the SQL queries solely based on the natural language question and the header embeddings.
We make another important assumption that the names given to the column headers are relevant to the information they are storing. If this is not the case, our model would not be able to perform well as it uses the header embeddings to generate a context for what the columns represent.
We have refrained from using Execution Guided Decoding to boost our performance, as it does not align with our concept of maintaining complete data privacy.
The loss of information to the model is evident and that is what impairs its performance. In our study, we have tried to show no compromise on the data privacy aspect of the tradeoff, whilst trying to maintain as high a performance as possible on the NL2SQL task.

\section{Future Prospects and Applications}
The model we have proposed currently predicts only a particular type of query, namely single table queries. In future works the model can be generalized to a great extent to support multi-table queries, multi column queries as well as a combination of the two. This would require the implementation of an SQL join prediction submodel as well as a select number prediction submodel as well as a larger and richer dataset containing various types of such queries to train the model on.  Presently the model only works with SELECT statements but many database users might require the UPDATE and DELETE functionality as well, which could be added to the model as well. WHERE isn’t the only SQL clause that is used to filter query results. The model can also be extended to support other clauses like HAVING and GROUP BY. Furthermore the concept of the NL2SQL task can be extended to support Data Definition Languages (DDL) with the purpose of creating databases and tables instead of just Data Manipulation Languages (DML) to query an existing database. The scope for NL2SQL remains vastly unexplored.

In our model we have employed the base model of RoBERTa (RoBERTa-base) to create embeddings of the natural language question and the headers. However, there exists an even larger model of RoBERTa, called RoBERTa-large, which has more parameters and has been proven to have higher accuracy in masked language modeling tasks and next sentence prediction tasks. Hence, by using RoBERTa-large to generate the embeddings which are then passed into the model, it may be possible to attain even higher accuracy on the training and test sets as well as the zero-shot learning task.

This model can be deployed as an Application Programming Interface (API) for use in fields where data privacy is of critical importance. This list includes, but is not limited to, military databases, medical record databases, financial information databases and market information databases. Since our model is data agnostic, it is also highly scalable, as it does not have to process all the cell data contained in the database. This makes it suitable for applications even where the database size is very large.

\section{Conclusion}
In this paper, we demonstrate the use of a data-agnostic model on a popular semantic parsing task, NL2SQL on the WikiSQL dataset. We observed that existing approaches either use table data as a part of the input features or for execution guided decoding. Since the primary focus of our work was on data privacy, we propose a completely data-blind model that effectively predicts SQL queries using only the table schema and natural language questions by generating knowledge vectors through word matching techniques. 

As described in the Results section, our model achieves 77.0 \% execution accuracy on the dev set and 76.7\% execution accuracy on the test set, on the WikiSQL dataset. This performance exceeds the latest content insentive approach viz TypeSQL, by a margin of almost 3\%. Creating a data blind model allows us to train a highly generalized model capable of zero shot learning.





%

\end{document}